# Neural Networks: According to the Principles of Grassmann Algebra

Z. Zarezadeh[1], N. Zarezadeh[2]
[1]Department of Electronic Engineering, University of Rome Tor Vergata, Via Cracovia, 90, 00133 Roma RM, Rome, Italy
[2] Department of Statistics, Mathematics and Computer, Allameh Tabataba'i University, Varzesh Sq., Dehkadeh Olampik, Tehran, Iran.

[1]E-mail: zakarya.zarezadeh@gmail.com
[2]E-mail: naeim.zarezade@gmail.com

**Abstract.** In this paper, we explore the algebra of quantum idempotents and the quantization of fermions which gives rise to a Hilbert space equal to the Grassmann algebra associated with the Lie algebra. Since idempotents carry representations of the algebra under consideration, they form algebraic varieties and smooth manifolds in the natural topology. In addition to the motivation of linking up mathematical physics with machine learning, it is also shown that by using idempotents and invariant subspace of the corresponding algebras, these representations encode and perhaps provide a probabilistic interpretation of reasoning and relational paths in geometrical terms.

*Introduction*: In recent decades, major efforts in the artificial neural network (ANN) community have focused on computational models of biological systems through the framework of interconnected networks of simple and often uniform units [1-5]. In connection with the theory of statistical mechanics and the spirit of the pioneering works [6-8] a wide class of simple mathematical models have been investigated from many perspectives and for different purposes.

Since then, ANNs have evolved into a broad family of techniques and have been successfully used to handle many complex and diverse tasks across multiple domains. The common feature of all these techniques is their structural similarity.

In recent years it has become obvious that the group theory and the representation theory of certain groups and algebras can contribute in an essential way to the understanding of how ANNs process information [9-13]. This strongly suggests that such representations should be of crucial importance and provide a natural guide for finding an effective learning theory. Recently a substantial amount of interest has been given to physics-inspired learning algorithms, such as quantum neural networks. A conglomeration of ideas and mathematical rigor to imitate or replace the classical neural networks with a quantum mechanical formalism. Within this framework, unitary gates and Hamiltonian operators are deployed to speculate the given dataset.

In this respect, motivated by the extension and limitations of the classical neural networks given briefly above, we will mainly focus on the generalization of theoretical formalism in the language of fermion algebra involving Grassmann directions or Grassmann variables. For clarity we provide only a description of the basic operations and transformation, however, the mathematical framework described in this paper allows for a direct generalization.

The structure of this paper is as follows: In Section 2 we define our notation and motivate the subject. Some essentials for our construction are given in Section 2. In Section 3 the formalism of the fermion algebra is introduced and the related group structure is discussed. We will describe Grassmann algebra and Clifford algebra which underlies symmetry and start to explore some of its matrix representations as associated with groups. The Section 4 contains conclusions.

Our standard references for the theory of modules, semisimple rings, and their representation are [14-15]. For Clifford algebras we use [16-18], on the representation theory of finite groups, we refer to [19-20] and for the group theory, we refer to [21-23].

*Grassmann algebra*: Grassmann algebra was first introduced by Hermann Grassmann in 1844 [24] and was rediscovered in the framework of quantum groups and supersymmetric quantum mechanics [25-27]. Interestingly the Grassmann algebras have found applications in representation theory, category of vector spaces, and physical theories about fermions and supersymmetry [28-30]. In this paper, we construct the irreducible representations of the Clifford algebras in terms of the Grassmann algebras. We explicitly show that the Grassmann generators give an axiomatic definition of abstract groups that encode a certain salient feature of the geometry. In particular, such an axiomatization formalizes the essential aspects of symmetries and provides a common mathematical framework to study neural network architectures and extends the remarkable potential of deep learning for a host of more complex applications.

A Grassmann algebra $\mathcal{A}$ on $\mathbb{R}$ or $\mathbb{C}$ (real or complex numbers) is an associative algebra generated by a unit (denoted by $I$ in what follows) and a set of generators $\{\theta_i\}$ that satisfy the anti-commutation relations

$$\theta_{ij} + \theta_{ji} = 0, \forall i,j$$

More precisely, given a set $\{e_1, e_2, ..., e_n\}$ with $n$ elements, Grassmann algebra of a vector space $E(n)$ over a field $\mathbb{k}$ is defined by the following exterior product $\wedge$ such that

$$e_i \wedge e_i = 0$$
$$e_i \wedge e_j = -e_j \wedge e_i$$

Together with the following properties:

$$(a \wedge b) \wedge c = a \wedge (b \wedge c) \; \forall a,b,c \in E(n)$$
$$a \wedge (\beta b + \gamma c) = \beta a \wedge b + \gamma a \wedge c, \forall a,b,c \in E(n), \beta, \gamma \in \mathbb{k}$$

The collection of the spaces $\wedge^k(V)$, k = 0,1,2,..., together with the operation $\wedge$ is called the exterior algebra on $V$. Where the $\wedge^k(V)$ is the subspace spanned by all vectors of the form

$$e_1 \wedge e_2 \wedge ... \wedge e_k$$

The set of all of the points in the Grassmann algebra is called the Grassmann directions or Grassmann variables and they are defined by their anti-commutation relations. The Grassmann variables are the basis vectors of a vector space. That is given two Grassmann numbers $\theta$ and $\partial$, the Grassmann algebra reads

$$\{\theta, \partial\} = \theta\partial + \partial\theta = 0$$

In particular, this means that given an anti-commuting number $\theta$ and $\partial$ they satisfy

$$\theta^2 = \partial^2 = 0$$

In terms of generalized Grassmann variables $\theta$ and its conjugate $\hat{\theta}$, we consider

$$\theta^k = \hat{\theta}^k = 0$$

where $k$ is a fixed number in $\mathbb{N}\setminus\{0,1\}$.(the particular case $k = 2$ corresponds to ordinary Grassmann variables). We then introduce derivative operators via

$$\partial_\theta f(\theta) = \frac{f(q\,\theta) - f(\theta)}{(q-1)\,\theta}$$

$$\partial_{\hat{\theta}} f(\hat{\theta}) = \frac{f(\hat{q}\,\hat{\theta}) - f(\hat{\theta})}{(\hat{q}-1)\,\hat{\theta}}$$

where $f$ and $g$ are arbitrary functions of $\theta$ and $\hat{\theta}$ and the complex number $q$ is chosen to be

$$q = \exp\left(\frac{2\pi i}{k}\right)$$

And $\hat{q}$ represents the complex conjugate of $q$. The linear operators $\partial_\theta$ and $\partial_{\hat{\theta}}$ satisfy

$$\partial_\theta \theta^n = [n]_q \theta^{n-1}$$

$$\partial_{\hat{\theta}} \hat{\theta}^n = [n]_{\hat{q}} \hat{\theta}^{n-1}$$

For $n = 0, 1, \ldots, k-1$, and the symbol $[\ ]_q$ is defined as

$$[X]_q = \frac{1-q^x}{1-q}, \quad \text{for } X \in \mathbb{R}$$

$$[n]_q = 1 + q + \cdots + q^{n-1}, \quad \text{for } n \in \mathbb{N}$$

Therefore, for functions $f: \theta \to f(\theta)$ and $g: \hat{\theta} \to g(\hat{\theta})$ one can define

$$f(\theta) = \sum_{n=0}^{k-1} \alpha_n \theta^n$$

$$g(\hat{\theta}) = \sum_{n=0}^{k-1} \beta_n \hat{\theta}^n$$

where $\alpha_n$ and $\beta_n$ are complex coefficients. In general, a Grassmann algebra on $n$ generators can be represented by $2^n \times 2^n$ square matrices. For our purposes, it is convenient to recall the Weyl construction [31-32] of the generators $\theta_i, \partial_i$ of the algebra

$$\theta_i = \underbrace{\omega \otimes \omega \otimes \ldots \otimes \omega}_{(i-1)\ term} \otimes \hat{\theta} \otimes \underbrace{I \otimes I \otimes \ldots \otimes I}_{(N-i)\ term}$$

$$\partial_i = \underbrace{\omega \otimes \omega \otimes \ldots \otimes \omega}_{(i-1)\ term} \otimes \hat{\partial} \otimes \underbrace{I \otimes I \otimes \ldots \otimes I}_{(N-i)\ term}$$

via the generators $\hat{\theta}$ and $\hat{\partial}$ of the 1-dimensional Grassmann algebra as

$$\hat{\theta}\hat{\partial} + \hat{\partial}\hat{\theta} = 1, \hat{\theta}^2 = 0 = \hat{\partial}^2, \hat{\omega} = [\hat{\partial}, \hat{\theta}]$$

with the corresponding matrix representation

$$\hat{\theta} = \begin{bmatrix} 0 & 0 \\ 1 & 0 \end{bmatrix}, \hat{\partial} = \begin{bmatrix} 0 & 1 \\ 0 & 0 \end{bmatrix}, \hat{\omega} = \begin{bmatrix} 1 & 0 \\ 0 & -1 \end{bmatrix}$$

For example, the $N$-dimensional Heisenberg-Weyl algebra is generated by $2N$ operators $\theta_i$ and $\partial_i$ satisfying the following commutation relation

$$\theta_i \theta_j + \theta_j \theta_i = 0$$
$$\partial_i \partial_j + \partial_j \partial_i = 0$$
$$\partial_i \theta_j + \theta_j \partial_i = \delta_{ij}$$

A particular application of these operators is in the quantum-mechanical treatment of angular momentum algebra and quantum harmonic oscillators [33]. In the context of the quantum theory, one can interpret these operators as creation and annihilation operators, adding or subtracting quanta of energy to the system. There are many far-reaching consequences of this algebraic theory of extended quantities. The most fundamental of these pertains to fermion algebra and supersymmetry.

***Exterior algebra construction of spinors***: Clifford algebra is the natural extension of Grassmann algebra on the bases of the simultaneous extension of the scalar and vector products [34]. A typical example is the algebra of multivectors with the geometric product as its operation

$$ab = a \cdot b + a \wedge b$$

which is the sum of the interior product $(a \cdot b)$ and exterior product $(a \wedge b)$. The general formula for the interior product between a vector v and a k-vector $w_1 \wedge w_2 \wedge \ldots \wedge w_k$ is as follows:

$$(W_1 \wedge W_2 \wedge \ldots \wedge W_k) \cdot v = \sum_{i=1}^{k} W_{i+1} \wedge W_{i+2} \wedge \ldots \wedge W_{i-1}(W_i \cdot v)$$

$$v \cdot (W_1 \wedge W_2 \wedge \ldots \wedge W_k) = \sum_{i=1}^{k} (v \cdot W_i) W_{i+1} \wedge W_{i+2} \wedge \ldots \wedge W_{i-1}$$

In general, the interior product is not associative and between a $k$-vector and a vector is commutative if $k$ is odd, and anticommutative if $k$ is even.

For a consistent description of spinor representation, a rather special role is played by Clifford algebra $Cl_{p,q}$ where $p$ indicates that $V$ has $p$-number of orthogonal basis with $e_i^2 = +1$ and $q$-number of orthogonal basis with $e_i^2 = -1$. (metric). Therefore, we can construct a Clifford algebra by specifying its basis vectors and their squares.

We will later on construct one such representation in terms of matrices, more specifically the operators

$$\theta^\mu + \eta^{\mu\nu} \frac{\partial}{\partial \theta^\nu}$$

acting on a Grassmann algebra would form one such representation, where $\eta^{\mu\nu} = \text{diag}(-1, \ldots, -1, +1, \ldots, +1)$ is the metric. To prepare for the discussion let us review the concept of idempotents. A projection or idempotent P on a vector space $v$ is a nonzero operator and a linear map $P: v \mapsto v$ such that $P^2 = P$, and two idempotents are orthogonal or annihilating if $P_+ P_- = P_- P_+ = 0$, hence $P_+$ projects onto one subspace and $P_- = (1 - P_+)$ projects onto the perpendicular subspace. Projectors can also be defined in terms of bra-ket notation as

$$P_i = |\phi_i\rangle\langle\phi_i|$$

where $|\phi\rangle$ denotes a vector in an abstract vector space $v$ and $\langle\phi|$ is covector to $|\phi\rangle$ which forms a subspace of the dual vector space.

There are different ways to combine vector spaces, for example, a two-level quantum system can be written as a combination of the basis vectors $\alpha|0\rangle + \beta|1\rangle$. Where $\alpha$ and $\beta$ are the contribution forms of each basis vector. Another way to combine vector spaces is via tensor product $|\psi\rangle \otimes |\phi\rangle$. If $|\psi\rangle = \sum_i \alpha_i |\psi_i\rangle$ and $|\phi\rangle = \sum_i \beta_i |\phi_i\rangle$, then the tensor product of these vectors can be written as

$$|\psi\rangle \otimes |\phi\rangle = \sum_{i,j} \alpha_i \beta_j |\psi_i\rangle \otimes |\phi_j\rangle = \sum_{i,j} \alpha_i \beta_j |\psi_i \phi_j\rangle$$

In this case, the projection operator

$$P_+ = |0\rangle\langle 0| = \begin{bmatrix} 1 & 0 \\ 0 & 0 \end{bmatrix}$$

Similarly

$$P_- = |1\rangle\langle 1| = (I - P_+) = \begin{bmatrix} 0 & 0 \\ 0 & 1 \end{bmatrix}$$

The main utility of the projection operators in this paper is the construction of anti-commuting elements of exterior algebra, nevertheless, they are used in quantum mechanics to determine the probability of measuring a particular observable for a given quantum state [35]. Consider, for example, the Grassmann algebra generated by two generators $\theta$ and $\partial$ by

$$\theta = (P_+ \otimes \theta_1) + (P_- \otimes \theta_1)$$
$$\partial = (\theta_1 \otimes P_+) - (\theta_1 \otimes P_-)$$

where $\theta_1 = |0\rangle \otimes |1\rangle$. Therefore, it is straightforward to obtain a realization of $2N$-dimensional Euclidean Clifford algebra $\{e_i, e_j\} = 2\delta_{ij}$ as following (where $i = 1, 2, \ldots, N$)

$$e_i = \partial_i + \theta_i$$
$$e_{N+i} = i(\partial_i - \theta_i)$$

In this way, one can construct the Clifford algebras, algebras of quantum idempotents and their representations. More precisely, Clifford algebras may be thought of as quantization (quantum deformation) of the Grassmann algebra [36].

Note that the action of $\theta$ and $\partial$ on the $k$-dimensional Hilbert space spanned by the orthonormal set $\{\,|\,n\,\rangle: n = 0, 1, \ldots, k-1\,\}$ is reminiscent of the finite quantum mechanics discussed by many authors [37-39]. However, another direction in which this formalism becomes very important is quantum computing in which the quantum idempotents generate Lie algebras to encode the dynamical symmetries of a quantum system.

One of the most important aspects of Clifford algebras is that they can be used to explicitly define the Lie algebra of the $Spin(n)$ group, whose underlying manifold is the double cover of the special orthogonal group $SO(n)$. A Lie group is a group and a smooth differentiable manifold whose group multiplication and inversion are smooth maps. For example, the $2 \times 2$ real invertible matrices form a general linear group $GL(2, \mathbb{R})$ of degree 2 under multiplication.

$$\mathrm{GL}(2, \mathbb{R}) = \{A = \begin{bmatrix} a & b \\ c & d \end{bmatrix} : \det(A) = a\,d - b\,c \neq 0\}$$

The group of $n \times n$ rotation matrices form a subgroup of $GL(n, \mathbb{R})$ denoted as $SO(n)$. The Lie algebra of $SO(n)$ is the space of skew-symmetric transformations of $\mathbb{R}^n$ with the map given by the elements of $\wedge^2(\mathbb{R}^n)$. A group of $n \times n$ invertible matrices with entries in $\mathbb{C}$ also form a Lie group denoted as $SU(n)$.

For example, the Lie algebra of $SO(3)$ denoted as $\mathcal{SO}(3)$, which consists of an antisymmetric $3 \times 3$ basis given by:

$$L_X = \begin{bmatrix} 0 & 0 & 0 \\ 0 & 0 & -1 \\ 0 & 1 & 0 \end{bmatrix}, L_Y = \begin{bmatrix} 0 & 0 & 1 \\ 0 & 0 & 0 \\ -1 & 0 & 0 \end{bmatrix}, L_Z = \begin{bmatrix} 0 & -1 & 0 \\ 1 & 0 & 0 \\ 0 & 0 & 0 \end{bmatrix}$$

with the following commutation relations

$$[L_X, L_Y] = L_Z$$
$$[L_Z, L_X] = L_Y$$
$$[L_Y, L_Z] = L_X$$

and one can use these to show that $1/2(e_i e_j)$ satisfies the same commutation relations

$$[\tfrac{1}{2}e_i e_j, \tfrac{1}{2}e_k e_l] = \delta_{il}(\tfrac{1}{2}e_k e_j) - \delta_{ik}(\tfrac{1}{2}e_l e_j) + \delta_{jl}(\tfrac{1}{2}e_i e_k) - \delta_{jk}(\tfrac{1}{2}e_i e_l)$$

Therefore, the $Spin(n)$ group can be constructed by exponentiating these quadratic elements of Clifford algebras by

$$e^{\theta\left(\frac{1}{2}e_i e_j\right)} = \cos\left(\frac{\theta}{2}\right) + e_i e_j \sin\left(\frac{\theta}{2}\right)$$

Given the Grassmann algebra on 2 generators we can identify the quaternion group structure on 3-sphere isomorphic to the groups $Spin(3)$ and $SU(2)$, by

$$I = \hat{\theta}\theta + \theta\hat{\theta}$$
$$\mathbb{i} = \theta - \hat{\theta}$$
$$\mathbb{j} = \partial - \hat{\partial}$$
$$\mathbb{k} = \mathbb{i}\,\mathbb{j}$$

where the set $\{\pm I, \pm\mathbb{i}, \pm\mathbb{j}, \pm\mathbb{k}\}$ are basis elements of the quaternion group and these Grassmann numbers can be represented by $4 \times 4$ matrices:

$$\theta = \begin{bmatrix} 0 & 0 & 0 & 0 \\ 1 & 0 & 0 & 0 \\ 0 & 0 & 0 & 0 \\ 0 & 0 & 1 & 0 \end{bmatrix}, \partial = \begin{bmatrix} 0 & 0 & 0 & 0 \\ 0 & 0 & 0 & 0 \\ 1 & 0 & 0 & 0 \\ 0 & -1 & 0 & 0 \end{bmatrix}$$

In this way, we arrive at the exterior algebra construction of spinors and we demonstrated how basic algebraic operations translate into physics language.

*Conclusion:* In this work, we have studied how Grassmann algebra and quantum idempotents can generate essentially all fundamental algebras in mathematical physics. We have given the representations of $SU(2)$ algebra of spin, $SO(4)$ algebra, and algebra of fermions in terms of Grassmann variables. This powerful axiomatization of such extended quantities gives rise to astonishingly rich mathematical theories which are relevant to many research areas in physics, statistical mechanics, and machine learning. The purpose of such an axiomatic basis is to clarify and reveal deep connections between different areas of mathematics, and the development of such axiomatization allows the proper formulation of the problems and hopefully offers new ideas and inspiration for harnessing the power of group theory and non-commutative algebras in machine learning.

*Declaration of conflicting interests:* The author(s) declared no potential conflicts of interest with respect to the research, authorship, and/or publication of this article'.

*Funding statement:* The author(s) received no financial support for the research, authorship, and/or publication of this article.